%% file: main.tex
\documentclass[10pt,twocolumn,letterpaper]{article}

\usepackage{iccv}              


\definecolor{iccvblue}{rgb}{0.21,0.49,0.74}
\usepackage[pagebackref,breaklinks,colorlinks,allcolors=iccvblue]{hyperref}
\usepackage{amsmath}
\usepackage[normalem]{ulem}
\usepackage{bm}
\useunder{\uline}{\ul}{}
\usepackage{multirow}
\usepackage{xcolor}

\usepackage[normalem]{ulem}
\usepackage{threeparttable}
\usepackage{pifont}
\def\paperID{6748} 
\def\confName{ICCV}
\def\confYear{2025}

\title{3D Gaussian Splatting Driven Multi-View Robust Physical Adversarial Camouflage Generation}

 \author{\normalsize \textbf{Tianrui Lou$^{1,2}$ \; Xiaojun Jia$^{3}$ \; Siyuan Liang$^{4}$ \; 
Jiawei Liang$^{1}$ \; Ming Zhang$^{5}$ \; Yanjun Xiao$^{6}$ \; Xiaochun Cao$^{1,2,*}$} \\
\small{\textit{$^{1}$Sun Yat-Sen University \quad $^{2}$Peng Cheng Laboratory \quad $^{3}$Nanyang Technological University \quad $^{4}$National University of Singapore}} \\
\small{\textit{$^{5}$National Key Laboratory of Science and Technology on Information System Security \quad $^{6}$Nsfocus}} \\
{\tt\small \{loutianrui, jiaxiaojunqaq, pandaliang521\}@gmail.com \quad liangjw57@mail2.sysu.edu.cn} \\
{\tt\small zm\_stiss@163.com \quad xiaoyanjun@nsfocus.com \quad caoxiaochun@mail.sysu.edu.cn}
}

\begin{document}
\maketitle
\input{sec/0_abstract}    
\input{sec/1_intro}

\input{sec/2_related}

\input{sec/3_background}

\input{sec/4_method}

\input{sec/5_experiments}

\input{sec/6_conclusion}


{
    \small
    \bibliographystyle{ieeenat_fullname}
    \bibliography{main}
}

\end{document}

%% file: sec/0_abstract.tex
\begin{abstract}
Physical adversarial attack methods expose the vulnerabilities of deep neural networks and pose a significant threat to safety-critical scenarios such as autonomous driving. 
Camouflage-based physical attack is a more promising approach compared to the patch-based attack, offering stronger adversarial effectiveness in complex physical environments. 
However, most prior work relies on mesh priors of the target object and virtual environments constructed by simulators, which are time-consuming to obtain and inevitably differ from the real world. 
Moreover, due to the limitations of the backgrounds in training images, previous methods often fail to produce multi-view robust adversarial camouflage and tend to fall into sub-optimal solutions.
Due to these reasons, prior work lacks adversarial effectiveness and robustness across diverse viewpoints and physical environments. 
We propose a physical attack framework based on 3D Gaussian Splatting (3DGS), named PGA, which provides rapid and precise reconstruction with few images, along with 
photo-realistic rendering capabilities. 
Our framework further enhances cross-view robustness and adversarial effectiveness by preventing mutual and self-occlusion among Gaussians and employing a min-max optimization approach that adjusts the imaging background of each viewpoint, helping the algorithm filter out non-robust adversarial features.
Extensive experiments validate the effectiveness and superiority of PGA. 
Our code is available at: \url{https://github.com/TRLou/PGA}.
\end{abstract}
\vspace{-4mm}

%% file: sec/1_intro.tex
\section{Introduction}
\begin{figure}[t]
  \centering
   \includegraphics[width=1\linewidth]{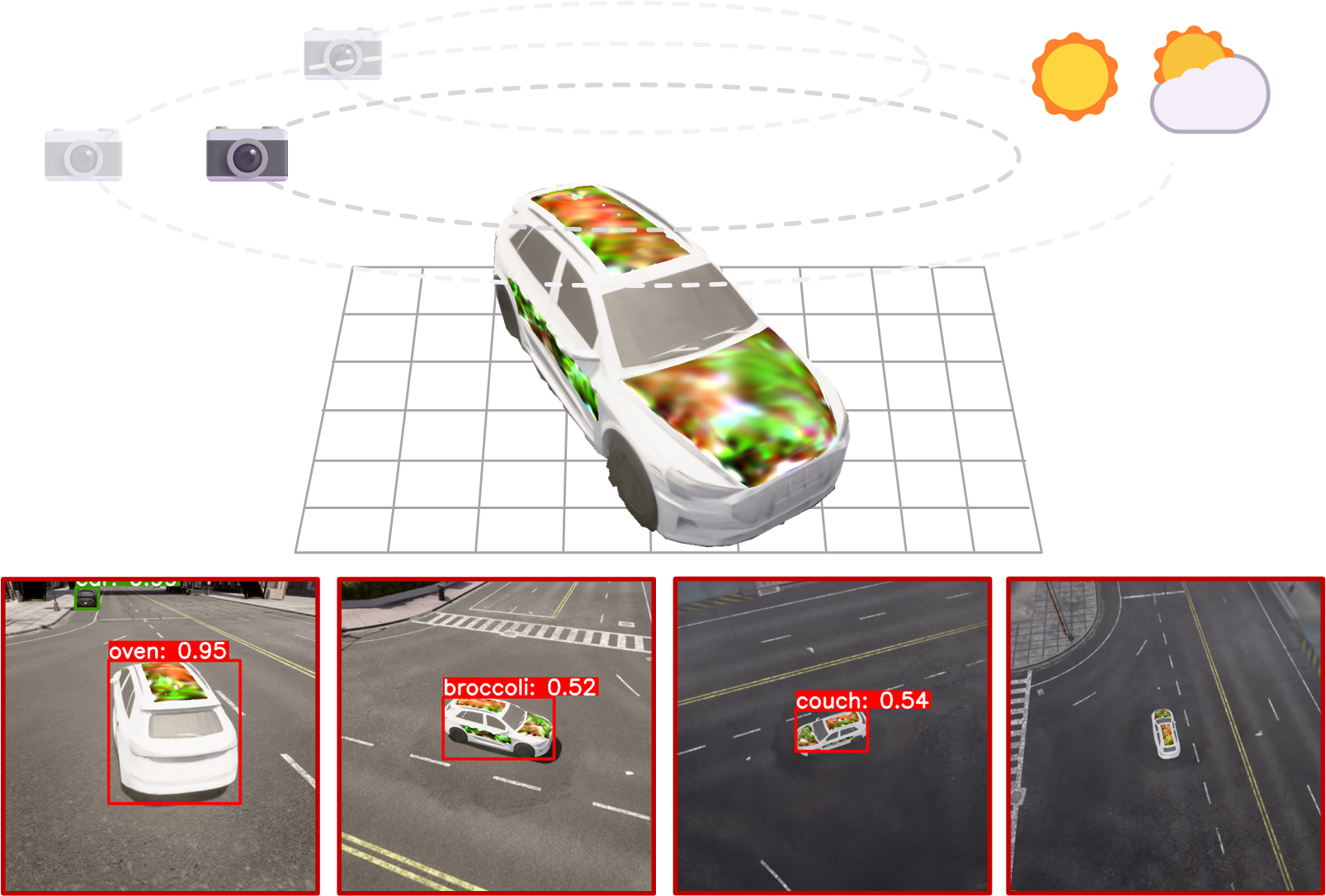}

   \caption{Visualization of multi-view robust adversarial camouflage generated by PGA, which effectively causes the victim detector to miss detections or misclassify the object across various environmental settings, including different shooting distances, pitch angles, azimuth angles, and weather conditions.
}
   \label{fig:teaser}
   \vspace{-4mm}
\end{figure}

\label{sec:intro}
Despite the remarkable success of deep neural networks (DNNs) in various fields, such as computer vision~\cite{he2016resnet} and natural language processing~\cite{vaswani2017attention, fan2023advances}, the emergence of adversarial attacks highlights the vulnerability of DNNs. 
Although digital attacks~\cite{goodfellow2014FGSM, gu2023survey, carlini2017cw,jia2020adv,he2023generating,lou2024hitadv,jia2025adversarial, muxue2023adversarial,jia2024semantic} targeting various tasks have raised concerns, physical attacks deployed in the real world pose even greater threats, stifling the use of DNNs in safety-critical domains such as autonomous driving~\cite{wang2023does, cao2023you, deng2020analysis}, security surveillance~\cite{nguyen2023physical, wang2019advpattern, liang2021generate, liang2020efficient, wei2018transferable, liang2022parallel, liang2022large, kong2024patch, liang2024object, kong2024environmental}, and remote sensing~\cite{wang2024fooling, lian2022benchmarking, liu2023x}.
We focus on physical attacks in autonomous driving, primarily targeting vehicle detection.


Physical attacks are often crafted in the digital domain and subsequently implemented by altering the physical properties of the target, such as patch application~\cite{eykholt2018rp2, brown2017adversarial, hu2021naturalistic} or camouflage deployment~\cite{wang2021DAS, wang2022FCA, wang2024TAS, suryanto2023ACTIVE, suryanto2022DTA, zhou2024rauca, zhang2018camou}.
The main challenge of physical attacks lies in minimizing the degradation of adversarial effectiveness when the generated adversarial camouflage is transferred from the digital domain to the physical domain, primarily due to environmental factors such as 
shooting distances, pitch angles, azimuth angles, and weather conditions.

Since adversarial camouflage offers higher robustness across different environmental settings compared to adversarial patches, it has become a more prevalent research direction.
Unlike adversarial patches, which only require pixel-level addition to the image of the target object during optimization, adversarial camouflage involves more complex shape-conforming computations.
Zhang et al.~\cite{zhang2018camou} and Wu et al.~\cite{wu2020genetic} estimate the image of camouflage applied to the target object through black-box methods, using a neural approximation function and a genetic algorithm, respectively.
Furthermore, to enhance adversarial effectiveness and robustness, a series of subsequent works~\cite{wang2021DAS, wang2022FCA, suryanto2022DTA, suryanto2023ACTIVE, wang2024TAS, zhou2024rauca} develop and employ differentiable neural renderers to render images based on the target object’s mesh. 
The differentiability of renderers allows for more precise white-box computation of adversarial camouflage.

Despite the success of previous methods to some extent, the robustness and adversarial effectiveness of the generated camouflage in physical environments remain limited due to the following two main reasons.
Firstly, these methods rely heavily on prior mesh information of the target object and virtual environments constructed by simulators such as CARLA~\cite{dosovitskiy2017carla}, which inevitably exhibit significant discrepancies from the real physical world.
Secondly, prior works usually apply only simple augmentations to backgrounds and viewpoints. 
The limited backgrounds in training images hinder the optimization of multi-view robust camouflage in the physical world, often resulting in sub-optimal solutions and leading to low robustness and universality.

In this paper, we propose a multi-view robust physical 3DGS-based attack method (PGA), which employs 3DGS as the differentiable rendering pipeline.
Thanks to the excellent reconstruction capabilities of 3DGS, PGA can quickly and accurately reconstruct the target object and background scene using only a few images, without the need for manually constructing.
Additionally, 3DGS enables fast, differentiable rendering from specified camera viewpoints, providing photo-realistic imaging results in the iterative attack process of PGA.
Furthermore, we propose to enhance the cross-view robustness and adversarial effectiveness of PGA by several methods.
Firstly, we address the issue of imaging inconsistency in adversarial camouflage across different viewpoints by preventing both mutual and self-occlusion among Gaussians.
Secondly, to generate physically adversarial camouflage that is robust and universal across various viewpoints, we design a min-max optimization approach. 
Concretely, we first add pixel-level perturbations to the background of each viewpoint's rendered image to maximize the detection loss, and then optimize the camouflage to minimize the loss, thereby obtaining multi-view robust adversarial features.
Finally, we incorporate several common techniques and regularization terms in the loss function to further enhance the physical performance and visual naturalness of the camouflage, including Expectation over Transformations (EoT)~\cite{athalye2018eot}, Non-Printability Score (NPS)~\cite{sharif2016nps} and primary color regularization.
Extensive experiments demonstrate that our attack framework outperforms state-of-the-art methods in both the digital and physical domains. 
Moreover, leveraging the features of 3DGS, our approach enables rapid modeling and effective attacks on various objects in autonomous driving scenarios and can be extended to attack tasks in infrared object detection, please refer to the supplementary material.

Our main contributions are in three aspects:
\begin{itemize}
\item We propose the first physical adversarial attack framework based on 3D Gaussian Splatting. 
Leveraging the precise and fast reconstruction capabilities of 3DGS, our PGA framework enables attacks on arbitrary objects in the physical world.



\item We further enhance the cross-view robustness and adversarial effectiveness. Firstly, we solve cross-view imaging inconsistency of camouflage by preventing mutual occlusion and self-occlusion of 3DGS.  Secondly, we propose a min-max optimization method to filter out multi-view non-robust adversarial features.

\item Extensive experiments validate the superiority of our framework to the state-of-the-art physical attack methods.

\end{itemize}

%% file: sec/2_related.tex
\section{Related Work}
\label{sec:related}
\paragraph{Physical Adversarial Attack.}
Most existing physical attack studies focus on autonomous driving scenarios, such as traffic sign detection~\cite{duan2020naturalstyles, feng2021metaattack, eykholt2018rp2, song2018physical_objectdetection}, pedestrian detection~\cite{sun2023differential, hu2022adversarial, hu2023physically, huang2020universal, xu2020adversarial, thys2019fooling}, and vehicle detection~\cite{zhang2018camou, wu2020genetic, wang2024TAS, zhang2023boosting, suryanto2022DTA, suryanto2023ACTIVE, wang2021DAS, wang2022FCA, zhu2024multiview}.
Compared to the previous two scenarios, physical attacks on vehicle detection are more challenging, as adversarial perturbations must remain robust across varying view angles, distances, and weather conditions.
Given that some studies have revealed the inadequacy of patch-based physical attacks in meeting the stringent robustness demands, researchers have opted to devise adversarial camouflage as an alternative.
To obtain camouflage and iteratively enhance its adversarial capability, a differentiable rendering process is essential.
As initial attempts, some studies employed black-box methods to estimate the rendering results.
Concretely, Zhang et al.~\cite{zhang2018camou} proposed to train a neural approximation function to imitate the rendering process,
and Wu et al.~\cite{wu2020genetic} computed optimal adversarial camouflage using a genetic algorithm.
To leverage a white-box setting for enhanced adversarial capabilities, some studies~\cite{wang2021DAS, wang2022FCA, suryanto2022DTA, suryanto2023ACTIVE, wang2024TAS} have focused on employing differentiable rendering method~\cite{Kato2018NMR, suryanto2022DTA}.
Wang et al.~\cite{wang2021DAS} proposed to suppress both model and human attention to gain visual naturalness and robustness.
Additionally, they later introduced further suppression of model-shared attention to enhance transferability~\cite{wang2024TAS}.
To overcome partial occluded and long-distance issues, Wang et al.~\cite{wang2022FCA} optimized full-coverage vehicle camouflage.
Suryanto et al.~\cite{suryanto2022DTA} designed a more photo-realistic renderer and integrated it into the attack framework, effectively enhancing the robustness of the camouflage.
Moreover, they improved robustness and universality by utilizing tri-planar mapping and making targets both misclassified and undetectable~\cite{suryanto2023ACTIVE}.
Zhou et al.~\cite{zhou2024rauca} addressed the complexities of weather conditions in physical scenarios by enhancing the neural renderer to accurately project vehicle textures and render images with environmental features like lighting and weather, forming the foundation of the RAUCA attack framework.

\paragraph{3D Modeling for Physical Attacks.}
Most of the above works require obtaining the mesh model of the target object in advance, which is time-consuming and labor-intensive.
Recently, some 3D representations have made it easier to model new objects and provide differentiable rendering pipelines that can be employed in physical attack frameworks, e.g. NeRF~\cite{mildenhall2021nerf}, 3D Gaussian Splatting~\cite{kerbl20233dgs}.
Li et al.~\cite{li2023adv3d} modeled target vehicles as NeRFs and optimized adversarial patches, resulting in improved physical realism.
Huang et al.~\cite{huang2024tt3d} proposed a transferable targeted attack approach that uses a grid-based NeRF to reconstruct the target object’s mesh, optimizing both texture and geometry simultaneously during iterations.
Despite these attack methods eliminating the dependency on the target object's mesh information, they are often limited by inherent drawbacks of NeRF, such as slow rendering, low quality, and high memory requirements.
In this paper, we resort to 3DGS, which can rapidly and accurately reconstruct the scene using numerous 3D Gaussian ellipsoids and easily perform differentiable, photo-realistic multi-view rendering, serving as the 3D representation of the target object to implement a physical attack framework.

%% file: sec/3_background.tex
\section{Preliminaries}
\label{sec:preli}
In this section, we will first provide a brief introduction to 3DGS.
Then we will analyze the challenges of generating deployable and effective adversarial camouflage using 3DGS as a differentiable rendering pipeline in the physical attack framework.

3DGS reconstructs the scene by representing it with a large set of Gaussians $\mathcal{G} = \{\bm{g}_1, \bm{g}_2, ..., \bm{g}_N\}$, where $N$ denotes the number of Gaussians.
Each Gaussian $\bm{g}$ is characterized by its mean $\mu_g$ and anisotropic covariance $\bm{\Sigma}_g$, and can be mathematically represented as:
\begin{equation}
\vspace{-1mm}
\bm{g}(\bm{x}) = exp(-\frac{1}{2}(\bm{x}-\mu_g)^T\bm{\Sigma}_g^{-1}(\bm{x}-\mu_g)),
\label{eq:}
\end{equation}
where the mean $\mu_g$ determines its central position, and the covariance $\bm{\Sigma}_g$ is defined by a scaling vector $\bm{s}_g \in \mathbb{R}^3$ and a quaternion $\bm{q}_g \in \mathbb{R}^4$ that encodes the rotation of $\bm{g}$.
Besides, 3DGS uses an $\alpha_g \in [0, 1]$ to represent the opacity of $\bm{g}$ and describes the view-dependent surface color $\bm{c}_g$ through spherical harmonics coefficients $\bm{k}_g$. 
To reconstruct a new scene, 3DGS requires only a few images $\mathcal{I}$ from different viewpoints as training inputs. 
Starting from a point cloud initialized by SfM~\cite{noah2006sfm}, it optimizes and adjusts the parameters $\{\mu_g, \bm{s}_g, \bm{q}_g, \alpha_g, \bm{k}_g\}$ of each $\bm{g}$ to make the rendering closely resemble the real images.
After training, an image $\bm{I_{\theta_c}}$ can be differentially rendered through a rasterizer $\mathcal{R}$ by splatting each 3D Gaussian $\bm{g}$ onto the image plane as a 2D Gaussian, with pixel values efficiently computed through alpha blending given a viewpoint $\bm{\theta_c}$ and a set $\mathcal{G}$, formulated as $\bm{I_{\theta_c}} = \mathcal{R}(\bm{\theta_c}, \mathcal{G})$. 
Then, the rendered images $\mathcal{I}_r = \{\bm{I_{\theta_{c1}}}, \bm{I_{\theta_{c2}}},...\}$ from various viewpoints are fed into the target detector $\mathcal{F}(\cdot;\bm{\theta_f})$, parameterized by $\bm{\theta_f}$, for evaluation.
The objective of our attack framework is to iteratively refine the attributes of the Gaussians $\mathcal{G}$ to mislead the detection results of $\mathcal{F}$, ultimately yielding robust adversarial Gaussians $\mathcal{G}'$ and camouflage $\bm{\mathcal{T}}$.

\begin{figure}[t]
  \centering
   \includegraphics[width=1\linewidth]{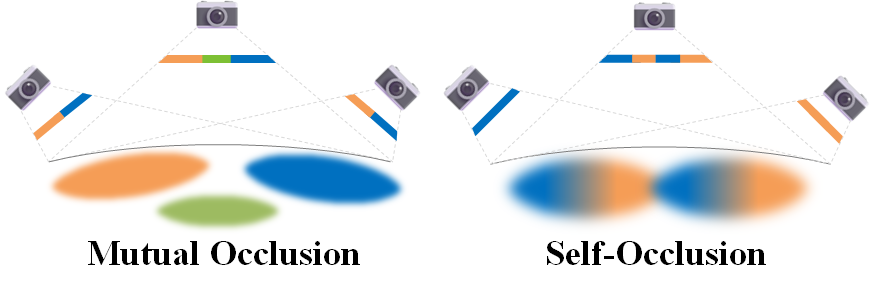}

   \caption{Illustration of mutual occlusion and self-occlusion issues in vanilla 3DGS that lead to cross-view inconsistencies.
}
   \label{fig:occlusion}
   \vspace{-4mm}
\end{figure}

\vspace{-4mm}
\paragraph{Problem Analysis}
We resort to 3DGS to support the proposed attack framework, which brings numerous advantages, including rapid reconstruction of arbitrary scenes and fast, differentiable rendering capabilities. However, generating camouflage with strong adversarial effectiveness and robustness in the physical world remains a challenge due to two main reasons.
\textbf{Firstly}, while vanilla 3DGS generally produces rendered images that align with the training set, discrepancies often exist between the represented 3D objects and their true values.
Concretely, not all Gaussians are positioned accurately on the surface, leading to mutual-occlusion issues among the Gaussians when the viewpoint changes. 
Additionally, since 3DGS uses spherical harmonics with strong representational capabilities to describe surface color, the same Gaussian may exhibit vastly different colors due to self-occlusion when the viewpoint changes. 
The issues of mutual occlusion and self-occlusion result in significant inconsistencies in the rendered camouflage across different viewpoints, reducing adversarial effectiveness and hindering physical deployment.
Please refer to Fig.~\ref{fig:occlusion}.
\textbf{Secondly}, 
in real-world scenarios, there are numerous factors affecting imaging results and detector performance, including shooting distance, angle, and weather conditions. 
During training, the limited variety of backgrounds makes it challenging to ensure that the generated adversarial camouflage is both universal and robust in real-world settings, leading traditional optimization methods to often fall into suboptimal solutions.
We address these two challenges individually and provide a detailed explanation in the following sections.

%% file: sec/4_method.tex
\section{Methodology}

\begin{figure*}[t]
  \centering
   \includegraphics[width=1\linewidth]{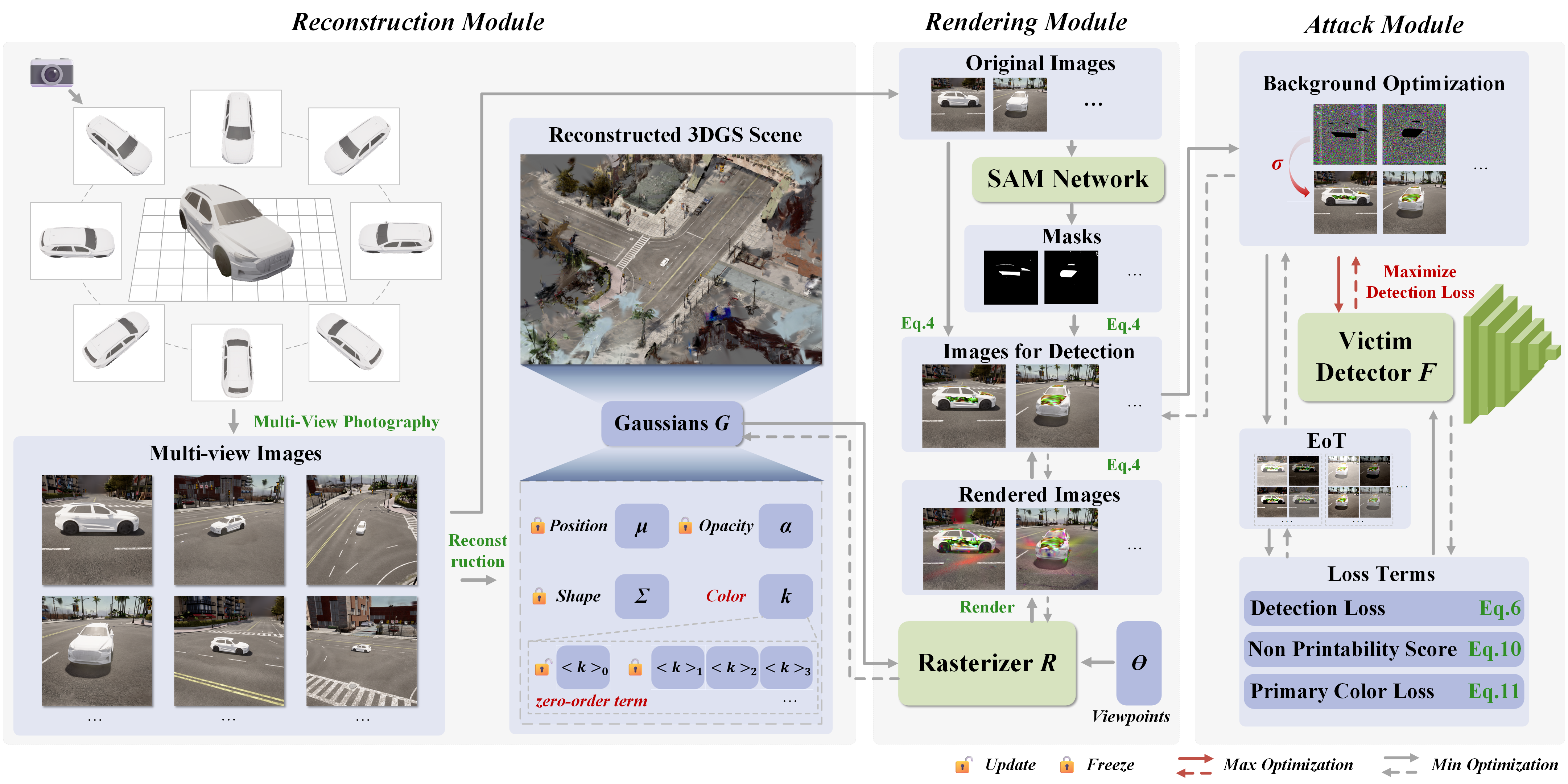}

   \caption{Demonstration of the framework of PGA. First, the reconstruction module captures multi-view images to build a 3DGS scene. Then the rendering module combines the clean background with the rendered adversarial camouflage to create the image for detection. Finally, the attack module applies a min-max optimization framework, first adding noise to the background to increase attack difficulty, then refining a multi-view robust camouflage with high adversarial effectiveness.
}
   \label{fig:framework}
   \vspace{-2mm}
\end{figure*}

We propose a novel physical attack framework based on 3D Gaussian Splatting named \textbf{PGA}.
We first introduce the pipeline and formulation of our framework in Sec.~\ref{sec:4.1}, then followed by the proposed strategies to enhance physical adversarial effectiveness and robustness in Sec.~\ref{sec:4.2}.

\subsection{Pipeline and Formulation of PGA Framework}
\label{sec:4.1}
The overall pipeline of our framework is shown in Fig~\ref{fig:framework}.
Our framework is composed of three components including a reconstruction module, a rendering module and an attack module.

\vspace{-4mm}
\paragraph{Reconstruction module.}
Given a set of images $\mathcal{I}=\{\bm{I_1}, \bm{I_2}, ...\}$ from different viewpoints, we first reconstruct the Gaussians $\mathcal{G}=\{\bm{g_1}, \bm{g_2}, ..., \bm{g_N}\}$ of  entire scene using the 3DGS training framework~\cite{kerbl20233dgs}.

\vspace{-4mm}
\paragraph{Rendering module.}
We select multiple camera viewpoints $\Theta = \{\bm{\theta_{c1}},\bm{\theta_{c2}},...\}$ around the target object at varying distances, pitch angles, and azimuth angles, ensuring comprehensive coverage to facilitate the generation of physically robust adversarial camouflage.
Then we obtain rendered images through rasterizer $\mathcal{R}$ provided by 3DGS:
\begin{equation}
\vspace{-1.5mm}
\mathcal{I}_r = \mathcal{R}(\Theta, \mathcal{G}).
\label{eq:}
\end{equation}
To ensure that adversarial perturbations are only added to the target object, we use SAM~\cite{kirillov2023SAM} to extract masks $\mathcal{M}$ from $\mathcal{I}_r$,
\begin{equation}
\vspace{-2mm}
\mathcal{M} = \text{SAM}(\mathcal{I}_r, \mathcal{P}),
\label{eq:}
\end{equation}
where $\mathcal{P}$ are prompts of the target object.
We create a copy of the original rendered image $\mathcal{I}_r$ as $\mathcal{I}_{ori}$, and the final images to be detected can be expressed as:
\begin{equation}
\vspace{-2mm}
\mathcal{I}_{\text{det}} = (\mathcal{I}_r \cdot \mathcal{M}) + (\mathcal{I}_{\text{ori}} \cdot (1 - \mathcal{M})) 
\label{eq:}
\end{equation}

\vspace{-4mm}
\paragraph{Attack module.}
After calculating $\mathcal{I}_{det}$ , we feed it into the victim object detector $\mathcal{F}$ to obtain the detection results:
\begin{equation}
\vspace{-2mm}
\mathcal{B} = \mathcal{F}(\mathcal{I}_{\text{det}}; \bm{\theta_f}) = \{\bm{b_{\theta_{c1}}}, \bm{b_{\theta_{c2}}},...\}.
\label{eq:}
\end{equation}
And the detection loss can be defined following~\cite{hu2023physically} as: 
\begin{equation}
\begin{aligned}
\vspace{-2mm}
\mathcal{L}_{\text{det}}(\mathcal{I}_{\text{det}}) &= \sum_{\bm{I}} \text{Conf}_{m^*}^{(\bm{I})}, \\m^*&=\underset{m}{\text{argmax}}
 \text{IoU}(\bm{gt^{(I)}},\bm{b^{(I)}}_m),
\label{eq:}
\end{aligned}
\end{equation}
where $\bm{I}$ is each input of a batch in $\mathcal{I}_{\text{det}}$, $\bm{b}_m$ is the \( m \)-th bounding box of detection results and $\text{Conf}$ is corresponding confidence.
$\mathcal{L}_{\text{det}}$ minimize the confidence score of the correct class in the box which has the maximum Intersection over Union (IoU) score with the ground truth $\bm{gt}$.

The optimization objective of the attack module can be formulated as:
\begin{equation}
\vspace{-2mm}
\mathcal{G}' = \arg\min_{\mathcal{G}} \mathcal{L}_{\text{det}} (\mathcal{I}_{\text{det}}(\bm{\theta_c}, \mathcal{G})) 
\label{eq:}
\end{equation}
Considering the difficulty and feasibility of manipulating the shape of the target object in the physical domain, we only optimize the spherical harmonics coefficients $k_g$ of the Gaussians $\mathcal{G}$, which represent the surface color, in an iterative attack process with a learning rate $\eta$:
\begin{equation}
\vspace{-2mm}
\bm{k}^{t+1} = \bm{k}^t + \eta \nabla_{\bm{k}}\mathcal{L}_{\text{det}}(\mathcal{I}_{\text{det}}).
\label{eq:}
\end{equation}

Upon completion of the iterative attack, the adversarial camouflage mesh $\bm{\mathcal{T}}$ can be derived from the optimized Gaussians $\mathcal{G}'$ following~\cite{guedon2024sugar} and deployed in the physical environment.

\subsection{Physical Adversarial Effectiveness and Robustness Enhancement }
\label{sec:4.2}


\subsubsection{Improving Cross-Viewpoint Consistency}

To tackle the issue of mutual occlusion, we facilitate the regularization terms from SuGaR~\cite{guedon2024sugar} 
in the reconstruction module,
aligning the Gaussians with the object surface and encouraging the Gaussians to reduce their opacity.
These terms prevent the Gaussians from being optimized inside the object, ensuring that their surface color is not occluded by other Gaussians on the surface when the viewpoint changes.

Additionally, we observe that higher-order spherical harmonics provide Gaussians with strong representational power for surface color, causing different parts of a single Gaussian to exhibit vastly different colors. 
When the viewpoint changes, these colors can occlude each other. 
This phenomenon becomes especially evident during multi-view joint iterative attack optimization, as the optimizer tends to focus on refining the visible portions of each Gaussian from each viewpoint, resulting in significant local color variations.
To address this self-occlusion problem, we propose optimizing only the zero-order term of the spherical harmonic coefficients $\langle \bm{k} \rangle_0$ during iterative attacks, ensuring uniform color changes across the surface of each Gaussian.
With these two improvements, we can ensure that the same adversarial camouflage is optimized consistently during cross-view iterative optimization. 

\subsubsection{Multi-view Robust Adversarial Camouflage Optimization Method}
Since it can be ragarded as Universal Adversarial Perturbation (UAP)~\cite{moosavi2017uap} problem and the attack difficulty varies significantly across different viewpoints, we iteratively optimize the camouflage for each viewpoint in sequence. 
To avoid over-optimization on easier viewpoints, which could increase the difficulty of optimizing other viewpoints, we set an iteration limit for each viewpoint. 
Once the camouflage successfully attacks a given viewpoint, we skip the remaining iterations and proceed to optimize the next viewpoint.

Additionally, since the adversarial effectiveness of camouflage is affected by the background context features, we conduct a ``counter adversarial attack'' on the background to make the adversarial features more robust to background variations.
Concretely, before each optimization iteration of the camouflage, we add point-wise noise $\bm{\sigma}$ to the background and optimize it iteratively using I-FGSM~\cite{kurakin2018ifgsm}.
Note that the optimization stops once the detector can correctly detect the target object or the iteration limit is reached, as excessive interference would make the camouflage difficult to optimize.
This process can be formulated as a min-max optimization problem:
\begin{equation}
\begin{aligned}
\vspace{-2mm}
\mathcal{G}' = \arg\min_{\mathcal{G}} \max_{\bm{\sigma}}  \mathcal{L}_{\text{det}} (\mathcal{I}_{\text{det}}(\bm{\theta_c}, \mathcal{G}) + &\bm{\sigma} \cdot (1 - \mathcal{M}) ) \\
&\text{s.t.} \ ||\bm{\sigma}||_\infty \leq \epsilon,
\label{eq:}
\end{aligned}
\end{equation}
where $\epsilon$ is a hyper-parameter denoting the budget of $\bm{\sigma}$.

\input{table/MainTable}

\subsubsection{Optimization Objective}
In addition to addressing the key issues mentioned above, we employ several additional techniques within the physical 3DGS-based attack framework to further improve its adversarial effectiveness and imperceptibility in real-world scenarios.
Firstly, we employ Expectation over Transformation (EoT)~\cite{athalye2018eot} in the optimization process, a technique widely used in various physical adversarial attack methods. Specifically, we apply a set of physical transformations, such as randomizing the scale, contrast, brightness, and adding noise, to enhance robustness.
Secondly, we introduce Non-Printability Score (NPS)~\cite{sharif2016nps} to mitigate fabrication error:
\begin{equation}
\begin{aligned}
\vspace{-2mm}
\text{NPS} = \sum_{\bm{\hat{p}} \in \mathcal{C}(\mathcal{I}_{\text{det}})} \prod_{\bm{p'} \in P} |\bm{\hat{p}} - \bm{p'}|,
\label{eq:}
\end{aligned}
\end{equation}
where $P$ is a set of printable colors and $\mathcal{C}(\mathcal{I}_{det})$ is a set of RGB triples used in $\mathcal{I}_{det}$.
Finally, to make the adversarial camouflage more imperceptible, we extract all background pixels from the training set, specifically $\mathcal{I}_{ori} \cdot (1 - \mathcal{M})$. 
Using K-means clustering, we group the background colors and select the top-k colors as the primary colors for the camouflage. 
During optimization, we add a regularization term to ensure that the camouflage remains close to the primary colors:
\begin{equation}
\begin{aligned}
\vspace{-2mm}
\mathcal{L}_{\text{clr}} = \frac{1}{|\Omega|} \sum_{(x,y)\in\Omega} \min_i ||\mathcal{I}_{\text{det}}(x,y)-\bm{c}_i||_2,
\label{eq:}
\end{aligned}
\end{equation}
where $(x, y)$ represents the position of pixel and $\Omega = \{(x,y)|\mathcal{M}(x, y) \textgreater 0\}$ is the set of pixel locations where the mask is non-zero.
Further, we also constrain the $L_2$ norm distance of the spherical harmonics coefficients before and after the attack to be as small as possible.
Thus, the overall loss can now be reformulated as:
\begin{equation}
\begin{aligned}
\vspace{-2mm}
\mathcal{L}_{\text{total}} &= \mathcal{L}_{\text{det}} (T(\mathcal{I}_{\text{det}}(\bm{\theta_c}, \mathcal{G}) + \bm{\sigma} \cdot (1 - \mathcal{M})) ) \\&+ \lambda ( \text{NPS} + \mathcal{L}_{\text{clr}} + ||\langle \bm{k} \rangle_0-\langle \bm{k} \rangle_0^\text{ori}||_2)
\label{eq:}
\end{aligned}
\end{equation}
where $T$ is transformations of EoT and $\lambda$ is hyper-parameter and $\langle \bm{k} \rangle_0^\text{ori}$ presents initial values before attack.
Meanwhile, the overall optimization objective and the iterative update process of the spherical harmonics coefficients can be reformulated separately as:

\begin{equation}
\vspace{-3mm}
\mathcal{G}' = \arg\min_{\mathcal{G}} \max_{\bm{\sigma}}\mathcal{L}_{\text{total}},
\label{eq:}
\end{equation}

\begin{equation}
\vspace{-3mm}
\langle \bm{k}^{t+1} \rangle_0 = \langle \bm{k}^{t} \rangle_0 + \eta \nabla_{\langle \bm{k} \rangle_0}\mathcal{L}_{\text{total}}.
\label{eq:}
\end{equation}

%% file: table/MainTable.tex
\begin{table*}[t]
\centering
\caption{Comparison results of AP@0.5(\%) for different physical attack methods on the COCO datasets targeting different detection models under different distances and weathers. Note that the adversarial camouflage is generated using Faster R-CNN and evaluated for black-box transferability on YOLO-v5, Mask R-CNN and Deformable-DETR.}
\vspace{-2mm}
\scalebox{0.78}{

\begin{tabular}{@{}c|c|cccc|cccc|c@{}}
\toprule
\multirow{2}{*}{Dis} & \multirow{2}{*}{Method} & \multicolumn{4}{c|}{Sunny}                                       & \multicolumn{4}{c|}{Cloudy}                                      & \multirow{2}{*}{Average} \\ \cmidrule(lr){3-10}
                     &                         & Faster R-CNN   & YOLO-V5*        & Mask R-CNN*      & D-DETR*         & Faster R-CNN   & YOLO-V5*        & Mask R-CNN*      & D-DETR*         &                          \\ \midrule
\multirow{8}{*}{5}   & -                       & 71.86         & 70.57          & 73.18          & 79.76          & 72.37         & 73.47          & 76.06          & 72.52          & 73.72                    \\
                     & DAS\cite{wang2021DAS}                     & 42.90         & 70.16          & 49.87          & 47.75          & 48.57         & 72.86          & 55.75          & 49.58          & 54.68                    \\
                     & FCA\cite{wang2022FCA}                     & 35.16         & 55.62          & 40.52          & 46.29          & 37.30         & 58.98          & 47.54          & 48.57          & 46.25                    \\
                     & DTA\cite{suryanto2022DTA}                     & 36.19         & 48.18          & 43.82          & 37.04          & 49.91         & 57.59          & 63.38          & 43.26          & 47.42                    \\
                     & ACTIVE\cite{suryanto2023ACTIVE}                  & 32.44         & 45.61          & 44.35          & 41.59          & 38.42         & 51.16          & 51.05          & 49.83          & 44.31                    \\
                     & TAS\cite{wang2024TAS}                     & 43.31         & 65.59          & 58.32          & 43.64          & 47.24         & 68.35          & 57.76          & 45.50          & 53.71                    \\
                     & RAUCA\cite{zhou2024rauca}                   & 21.71         & 46.94          & 31.90          & 36.54          & 27.85         & 56.01          & 36.50          & 39.79          & 37.16                    \\
                     & PGA                     & \textbf{4.52} & \textbf{39.10} & \textbf{10.62} & \textbf{28.31} & \textbf{5.60} & \textbf{46.99} & \textbf{16.67} & \textbf{35.90} & \textbf{23.46}           \\ \midrule
\multirow{8}{*}{10}  & -                       & 89.03         & 91.87          & 91.41          & 81.47          & 87.10         & 94.91          & 90.65          & 82.04          & 88.56                    \\
                     & DAS\cite{wang2021DAS}                     & 77.98         & 77.69          & 87.31          & 72.60          & 64.83         & 73.43          & 70.98          & 74.02          & 74.86                    \\
                     & FCA\cite{wang2022FCA}                     & 59.98         & 67.87          & 65.00          & 67.47          & 55.88         & 65.23          & 55.25          & 63.43          & 62.51                    \\
                     & DTA\cite{suryanto2022DTA}                     & 55.61         & 66.27          & 74.81          & 53.83          & 55.38         & 62.01          & 74.66          & 57.75          & 62.54                    \\
                     & ACTIVE\cite{suryanto2023ACTIVE}                  & 59.00         & 68.94          & 71.67          & 52.41          & 60.02         & 69.81          & 64.46          & 61,79          & 63.76                    \\
                     & TAS\cite{wang2024TAS}                     & 53.85         & 69.41          & 80.56          & 55.21          & 53.86         & 75.57          & 68.34          & 52.25          & 63.63                    \\
                     & RAUCA\cite{zhou2024rauca}                   & 18.88         & 56.70          & 31.00          & 44.85          & 21.74         & 59.37          & 34.29          & 47.17          & 39.25                    \\
                     & PGA                     & \textbf{1.40} & \textbf{45.53} & \textbf{8.44}  & \textbf{30.89} & \textbf{0.71} & \textbf{48.18} & \textbf{8.53}  & \textbf{30.54} & \textbf{21.78}           \\ \midrule
\multirow{8}{*}{15}  & -                       & 84.12         & 97.78          & 94.54          & 79.66          & 88.10         & 97.78          & 93.52          & 83.90          & 89.93                    \\
                     & DAS\cite{wang2021DAS}                     & 78.67         & 89.86          & 81.57          & 73.88          & 62.13         & 75.28          & 70.94          & 74.10          & 75.80                    \\
                     & FCA\cite{wang2022FCA}                     & 66.37         & 77.80          & 76.58          & 69.56          & 61.97         & 69.74          & 69.07          & 73.05          & 70.52                    \\
                     & DTA\cite{suryanto2022DTA}                     & 57.17         & 72.47          & 73.78          & 61.65          & 55.17         & 64.94          & 65.60          & 66.65          & 64.68                    \\
                     & ACTIVE\cite{suryanto2023ACTIVE}                  & 53.58         & 78.98          & 60.16          & 60.54          & 57.56         & 68.40          & 58.77          & 69.50          & 63.44                    \\
                     & TAS\cite{wang2024TAS}                     & 55.79         & 70.57          & 67.21          & 67.25          & 65.23         & 68.34          & 73.32          & 68.28          & 67.00                    \\
                     & RAUCA\cite{zhou2024rauca}                   & 37.80         & 63.32          & 58.27          & 44.69          & 38.46         & 64.97          & 46.19          & 56.73          & 51.30                    \\
                     & PGA                     & \textbf{1.95} & \textbf{52.96} & \textbf{9.40}  & \textbf{29.58} & \textbf{7.16} & \textbf{59.86} & \textbf{12.24} & \textbf{31.10} & \textbf{25.53}           \\ \midrule
\multirow{8}{*}{20}  & -                       & 86.50         & 96.81          & 91.99          & 83.37          & 86.60         & 98.89          & 92.35          & 85.08          & 90.20                    \\
                     & DAS\cite{wang2021DAS}                     & 68.67         & 88.47          & 78.52          & 76.14          & 60.62         & 69.47          & 65.95          & 70.69          & 72.32                    \\
                     & FCA\cite{wang2022FCA}                     & 64.23         & 71.53          & 78.88          & 72.99          & 58.87         & 63.60          & 66.96          & 73.72          & 68.85                    \\
                     & DTA\cite{suryanto2022DTA}                     & 48.99         & 76.44          & 74.89          & 70.40          & 58.14         & 65.48          & 70.14          & 68.99          & 66.68                    \\
                     & ACTIVE\cite{suryanto2023ACTIVE}                  & 39.70         & 70.77          & 64.31          & 67.28          & 50.47         & 65.02          & 57.00          & 70.66          & 60.65                    \\
                     & TAS\cite{wang2024TAS}                     & 67.20         & 84.92          & 85.21          & 70.25          & 57.33         & 74.42          & 71.13          & 64.54          & 71.88                    \\
                     & RAUCA\cite{zhou2024rauca}                   & 37.29         & 59.34          & 59.07          & 48.60          & 32.84         & 55.57          & 42.89          & 60.39          & 49.50                    \\
                     & PGA                     & \textbf{1.85} & \textbf{43.95} & \textbf{14.60} & \textbf{23.14} & \textbf{5.40} & \textbf{41.42} & \textbf{14.63} & \textbf{20.83} & \textbf{20.73}           \\ \bottomrule
\end{tabular}
}

\vspace{-4mm}
\label{tab:main}
\end{table*}

%% file: sec/5_experiments.tex
\begin{figure*}[t]
  \centering
   \includegraphics[width=1\linewidth]{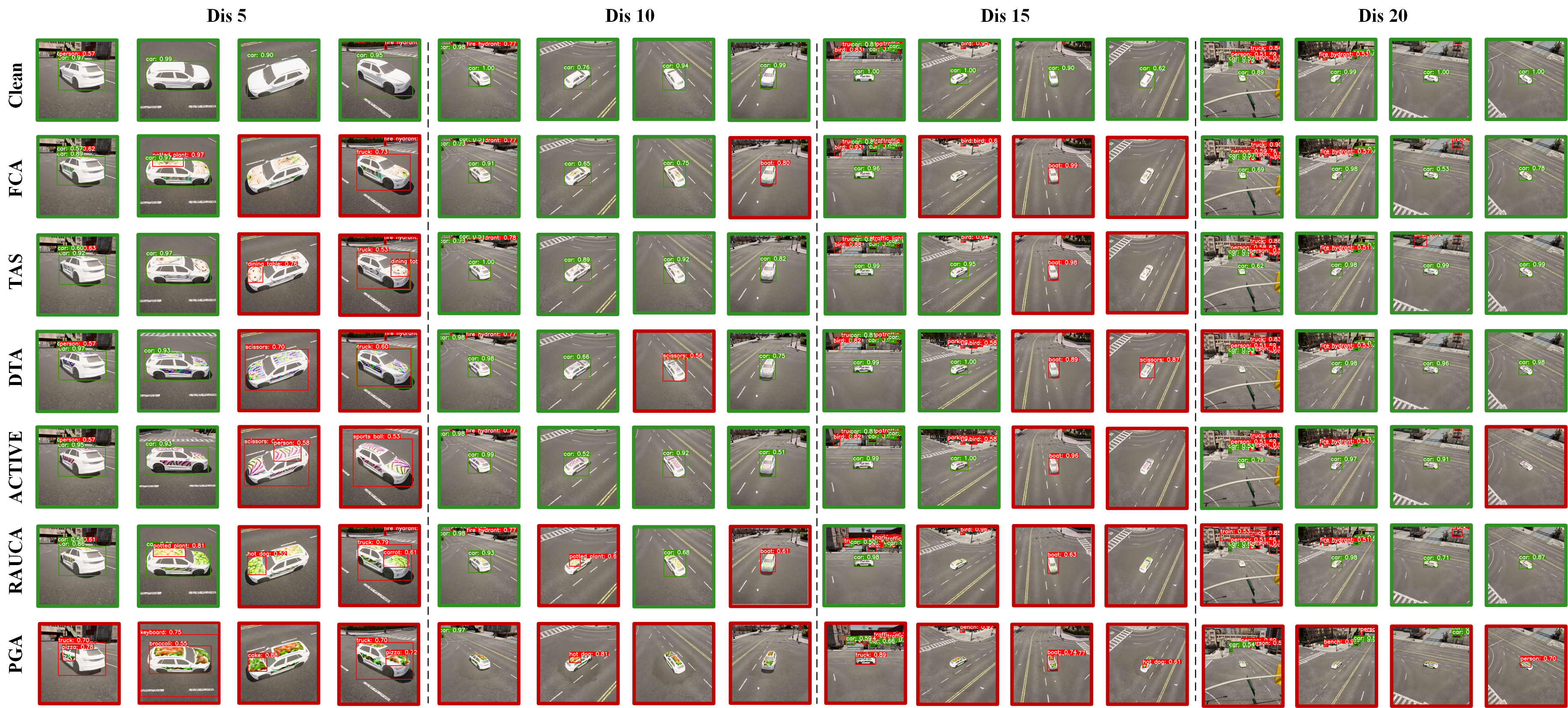}

   \caption{Visualization comparison of multi-view detection results in the digital world. Green-bordered images indicate correct detection of the target vehicle, while red-bordered images indicate either undetected targets or detection with incorrect classification.
}
   \label{fig:visual}
   \vspace{-4mm}
\end{figure*}
\section{Experiments}

In this section, we first illustrate the experimental settings and implementation details. 
We then demonstrate the superiority and effectiveness of PGA through digital domain experiments, including extensive qualitative and quantitative comparisons of attack performance 
and ablation studies. Furthermore, we conduct physical domain experiments, presenting outstanding qualitative and quantitative results of the generated camouflage on a 1:24 scale toy car and a 1:1 scale real vehicle.

\subsection{Experimental Setup}
\paragraph{Datasets.} 
To comprehensively validate the effectiveness of our attack method, we construct datasets for both the digital domain and the physical domain.

For the \textbf{digital domain dataset}, we use the CARLA simulation environment~\cite{dosovitskiy2017carla} based on Unreal Engine 4, a popular open-source simulator for autonomous driving scenarios, to construct high-fidelity and photo-realistic urban scenes. 
The test set for each attack method is created by capturing images with a camera positioned around the vehicle deployed with the corresponding adversarial camouflage. 
We select two kinds of weather (sunny and cloudy), four distances ($5m, 10m, 15m, 20m$) and five camera pitch angles ($20^\circ, 30^\circ, 40^\circ, 50^\circ, 60^\circ$) because COCO-pretrained detection models inherently have poor performance at greater distances or larger pitch angles, making those settings less informative for evaluation.
For each setting, we conduct $360^\circ$ surrounding photography at $10^\circ$ intervals
, resulting in 1440 images totally.
For the \textbf{physical domain dataset}, 
we deploy a 1:1 scale real vehicle, GOLF Sportsvan, then capture a rotating video using a drone, and extract 282 images to input into the PGA framework to generate camouflage. The camouflage is deployed using stickers.
Subsequently, we employ a drone to record videos and extract images analogous to the digital domain dataset to construct a test dataset.
We further deploy the adversarial camouflage of PGA and other SOTA methods on a 1:24 scale Audi Q5 model car to conduct additional qualitative and quantitative experiments across diverse scenarios.

\vspace{-5mm}
\paragraph{Target Models.}
We select commonly used detection model architectures for the experiments, including one-stage detectors: YOLO-v5; two-stage detectors: Faster R-CNN~\cite{ren2015frcnn} and Mask R-CNN~\cite{he2017maskrcnn}; as well as transformer-based detectors: Deformable-DETR~\cite{zhu2020ddetr}, with all models pre-trained on the COCO dataset.

\vspace{-5mm}
\paragraph{Compared Methods.} 
We select six state-of-the-art physical adversarial attack methods as our baseline for comparison, including DAS~\cite{wang2021DAS}, FCA~\cite{wang2022FCA}, DTA~\cite{suryanto2022DTA}, ACTIVE~\cite{suryanto2023ACTIVE}, TAS~\cite{wang2024TAS} and RAUCA~\cite{zhou2024rauca}.

\vspace{-5mm}
\paragraph{Evaluation Metrics.} 
To evaluate the effectiveness of various attack methods on detection models, we use AP@0.5($\%$), following~\cite{suryanto2022DTA, suryanto2023ACTIVE, zhou2024rauca}, which is a standard measure capturing both recall and precision at a detection IoU threshold of 0.5.

\input{table/angle}

\input{table/Ablation}

\subsection{Digital Experiments}
In this section, we provide a comprehensive comparison of PGA and SOTA methods, demonstrating the advantages of PGA. 
In these experiments, Faster R-CNN is used as the victim model for white-box attacks, with the adversarial camouflage transferred to other three detectors (marked with *, and * represents the same meaning throughout) to evaluate transferability. 
Note that we primarily use partial coverage camouflage in this section, following~\cite{wang2021DAS, wang2024TAS}. This setting is more challenging due to the reduced optimization space, yet we adopt it because it greatly facilitates real-world deployment and significantly lowers deployment costs. Additionally, we provide comparative experiments using full-coverage camouflage, where PGA still outperforms other methods; please refer to the Appendix.

\vspace{-5mm}
\paragraph{Digital World Attack.} 
We compare the digital attack performance of PGA with SOTA methods across multiple weather conditions, distances, and viewpoints. 
Although PGA can directly reconstruct and attack using real photos, for a fair comparison with mainstream vehicle physical attack methods, we sample clean vehicle images in CARLA, reconstruct the 3D scene, and then conduct the PGA attack.
The results in Tab.~\ref{tab:main} show that PGA achieves the best attack performance in all settings, indicating that the generated adversarial camouflage possesses high adversarial strength, high multi-view robustness, and strong transferability.

In addition, we conduct comparative experiments on camera pitch angles in Tab.~\ref{tab:angle}, and the results indicate that PGA consistently outperforms at all angles. We select views ranging from $20^\circ$ to $60^\circ$ because detectors pre-trained on COCO perform poorly at higher bird's-eye view angles.


\vspace{-4mm}
\paragraph{Visualization.} 
We present visualizations in Fig.~\ref{fig:visual} comparing the detection results of PGA with other attack methods in the digital domain.
The results indicate that, compared to other SOTA methods, PGA exhibits superior adversarial effectiveness and multi-view robustness across various distances, pitch angles and azimuth angles.
More results under different lighting and weather conditions are provided in the supplementary material.

\vspace{-4mm}
\paragraph{Ablation Study.}
We conduct an ablation experiment focusing on the two techniques in PGA, including multi-view camouflage consistency (Cons.) and the min-max optimization framework, with the results shown in Tab.~\ref{tab:ablation}. It is apparent that using all two techniques simultaneously achieves the best attack performance.
More ablation experiments are provided in the supplementary materials.

\subsection{Physical Experiments}

\begin{figure}[t]

   \includegraphics[width=1\linewidth]{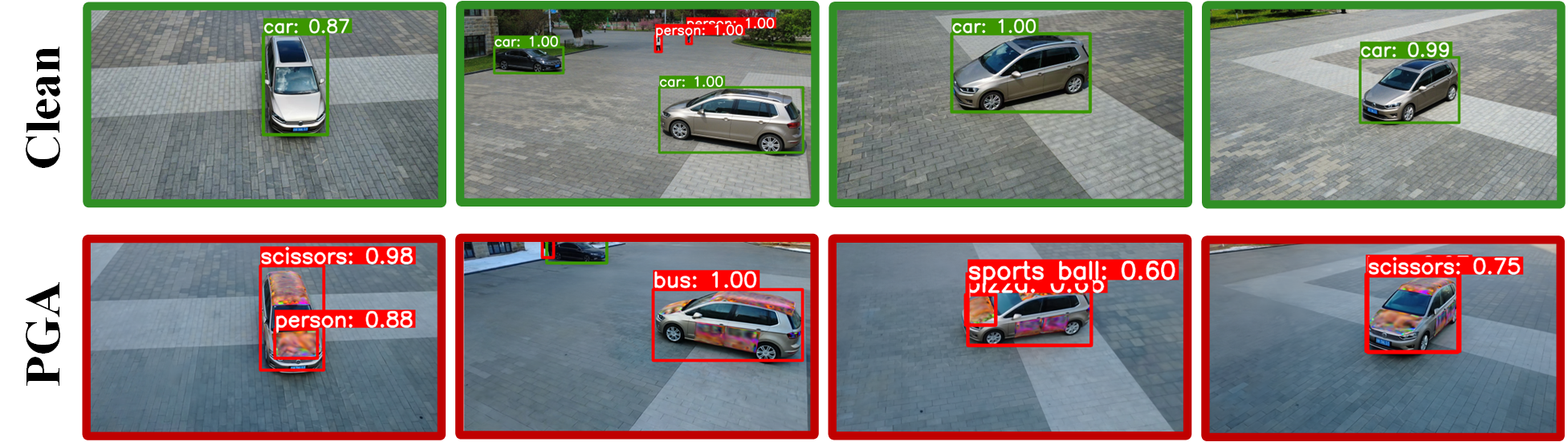}

   \caption{Visualization results from physical experiments on a 1:1 real car. We deploy the PGA adversarial camouflage using stickers and capture images from multiple viewpoints with a drone.
}
   \label{fig:realcar}
   \vspace{-4mm}
\end{figure}

\begin{figure}[t]
\centering
   \includegraphics[width=1\linewidth]{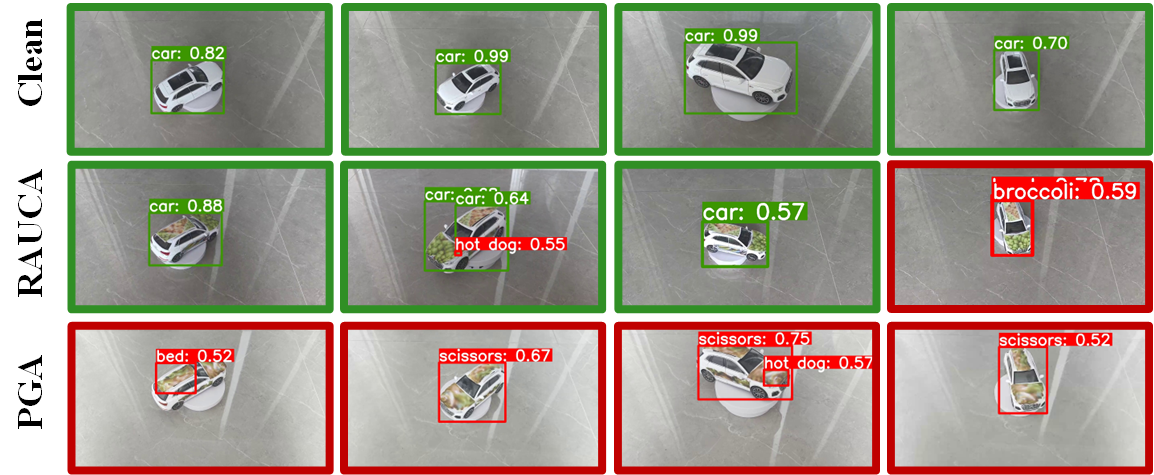}

   \caption{Visualization results from physical experiments on a 1:24 scale simulated car. We compare the attack visualization outcomes of clean samples, RAUCA~\cite{zhou2024rauca} camouflage samples, and PGA camouflage samples from multiple viewpoints.
}
   \label{fig:phy}
   \vspace{-2mm}
\end{figure}

\input{table/PhysicalTable}

\paragraph{1:24 Physical Experiment.}
We deploy adversarial camouflage generated by various SOTA methods as well as PGA on a 1:24 scale toy car. 
Images are captured from multiple viewpoints at distances of 50cm and 100cm to construct a physical scene dataset, which is subsequently evaluated using multiple detectors.
Quantitative results are provided in Tab.~\ref{tab:phy}, and qualitative results are shown in Fig.~\ref{fig:phy}. 
These results indicate that PGA's photo-realistic modeling capability and multi-view adversarial robustness can effectively simulate physical environments and mitigate the degradation in adversarial camouflage performance caused by these conditions.

\vspace{-4mm}
\paragraph{1:1 Physical Experiment.}
We also apply the PGA framework for 3DGS modeling and adversarial camouflage generation on a 1:1 real vehicle. 
During drone-based image capture, easy-to-deploy calibration stickers are used to help SAM segment the camouflage areas. 
We conduct attack on Faster R-CNN, where AP@0.5(\%) decreases from 88.48 to 25.67, with qualitative results presented in Fig.~\ref{fig:realcar}.
Results show that with little manual effort and simple tools (a camera and some printed stickers), PGA can effectively reconstruct and attack real cars, posing a significant threat to autonomous driving safety.

%% file: table/angle.tex
\begin{table}[]
\centering
\caption{Comparison of detection results for different attack methods at various pitch angles, specifically reporting the average AP@0.5 on Faster R-CNN for distances from 5m to 20m under both sunny and cloudy weather conditions.}
\vspace{-2mm}
\scalebox{0.65}{
\begin{tabular}{@{}c|ccccc|c@{}}
\toprule
Angle  & $20^\circ$             & $30^\circ$           & $40^\circ$            & $50^\circ$            & $60^\circ$            & Average       \\ \midrule
-      & 91.30          & 87.00         & 88.04         & 78.70         & 65.46         & 82.10         \\
FCA\cite{wang2022FCA}    & 61.19          & 60.61         & 65.34         & 52.84         & 27.54         & 53.50         \\
DTA\cite{suryanto2022DTA}    & 65.11          & 62.37         & 57.71         & 53.09         & 25.72         & 52.80         \\
ACTIVE\cite{suryanto2023ACTIVE} & 56.94          & 60.70         & 65.27         & 43.64         & 11.52         & 47.61         \\
RAUCA\cite{zhou2024rauca}  & 46.36          & 43.69         & 46.72         & 23.47         & 9.63          & 33.97         \\
PGA    & \textbf{21.01} & \textbf{4.62} & \textbf{4.11} & \textbf{3.90} & \textbf{0.00} & \textbf{6.73} \\ \bottomrule
\end{tabular}
}
\vspace{-2mm}

\label{tab:angle}
\end{table}

%% file: table/Ablation.tex
\begin{table}[]
\centering
\caption{Ablation study results of various techniques applied in the PGA attack framework.}
\vspace{-2mm}
\scalebox{0.65}{

\begin{tabular}{@{}cc|cccc|c@{}}
\toprule
Cons.                     & Min-Max                   & Faster R-CNN  & Yolo-v5*       & Mask R-CNN*    & D-DETR*        & Average        \\ \midrule
                          & \checkmark & 8.05          & 50.38          & 16.33          & 34.50          & 27.32          \\
\checkmark &                           & 10.23         & 54.40          & 20.56          & 36.82          & 30.50          \\
\checkmark & \checkmark & \textbf{3.57} & \textbf{47.24} & \textbf{11.89} & \textbf{28.78} & \textbf{22.87} \\ \bottomrule
\end{tabular}
}

\vspace{-4mm}
\label{tab:ablation}
\end{table}

%% file: table/PhysicalTable.tex
\begin{table}[t]
\centering
\caption{Comparison results of AP@0.5(\%) under physical settings. We deploy adversarial textures generated by different attack methods on a 1:24 scale toy car and capture images from multiple viewpoints at distances of 50cm and 100cm to construct a physical scene dataset for detection.}
\scalebox{0.6}{
\begin{tabular}{@{}c|c|cccc|c@{}}
\toprule
Dis                    & Method & Faster R-CNN   & YOLO-v5*        & Mask R-CNN*     & D-DETR*         & Average        \\ \midrule
\multirow{6}{*}{50cm}  & -      & 86.12          & 90.71          & 85.36          & 89.25          & 87.86          \\
                       & FCA\cite{wang2022FCA}    & 66.41          & 61.37          & 58.55          & 59.43          & 61.44          \\
                       & DTA\cite{suryanto2022DTA}    & 55.58          & 57.49          & 56.12          & 60.98          & 58.79          \\
                       & ACTIVE\cite{suryanto2023ACTIVE} & 39.45          & 52.38          & 47.31          & 45.95          & 46.27          \\
                       & RAUCA\cite{zhou2024rauca}  & 28.86          & 50.67          & 32.09          & 35.14          & 36.69          \\
                       & PGA    & \textbf{20.94} & \textbf{50.25} & \textbf{22.35} & \textbf{21.25} & \textbf{28.69} \\ \midrule
\multirow{6}{*}{100cm} & -      & 90.19          & 92.95          & 89.32          & 93.02          & 91.37          \\
                       & FCA\cite{wang2022FCA}    & 44.16          & 48.95          & 49.08          & 50.24          & 48.10          \\
                       & DTA\cite{suryanto2022DTA}    & 50.81          & 48.11          & 53.02          & 51.81          & 50.93          \\
                       & ACTIVE\cite{suryanto2023ACTIVE} & 40.10          & 52.35          & 45.39          & 49.28          & 46.78          \\
                       & RAUCA\cite{zhou2024rauca}  & 34.61          & 44.14          & 35.55          & 34.70          & 37.25          \\
                       & PGA    & \textbf{21.77} & \textbf{41.82} & \textbf{23.92} & \textbf{25.54} & \textbf{28.26} \\ \bottomrule
\end{tabular}
}

\vspace{-4mm}

\label{tab:phy}
\end{table}

%% file: sec/6_conclusion.tex
\section{Conclusion}
In this paper, we propose a novel physical attack framework based on 3D Gaussian Splatting named PGA.
Further, we improve the physical adversarial effectiveness and multi-view robustness by improving the cross-viewpoint consistency of the camouflage and using a multi-view robust min-max adversarial camouflage optimization method.
Experiments prove that PGA can effectively attack arbitrary objects in both digital and physical domains, even in infrared modality. 
We hope our work can inspire efforts to improve true robustness in the physical world.

\paragraph{Acknowledgment.}
Supported by \ding{172}Shenzhen Science and Technology Program(KJZD20240903095730039), \ding{173}the CCF-NSFOCUS `Kunpeng' Research Fund(CCF-NSFOCUS 2024003), \ding{174}Shenzhen Science and Technology Program(JCYJ20210324102204012) and \ding{175}the Fundamental Research Funds for the Central Universities, Sun Yat-sen University under Grants No. 23xkjc010.

%% file: main.bbl
\begin{thebibliography}{66}
\providecommand{\natexlab}[1]{#1}
\providecommand{\url}[1]{\texttt{#1}}
\expandafter\ifx\csname urlstyle\endcsname\relax
  \providecommand{\doi}[1]{doi: #1}\else
  \providecommand{\doi}{doi: \begingroup \urlstyle{rm}\Url}\fi

\bibitem[Athalye et~al.(2018)Athalye, Engstrom, Ilyas, and Kwok]{athalye2018eot}
Anish Athalye, Logan Engstrom, Andrew Ilyas, and Kevin Kwok.
\newblock Synthesizing robust adversarial examples.
\newblock In \emph{International conference on machine learning}, pages 284--293. PMLR, 2018.

\bibitem[Brown et~al.(2017)Brown, Man{\'e}, Roy, Abadi, and Gilmer]{brown2017adversarial}
Tom~B Brown, Dandelion Man{\'e}, Aurko Roy, Mart{\'\i}n Abadi, and Justin Gilmer.
\newblock Adversarial patch.
\newblock \emph{arXiv preprint arXiv:1712.09665}, 2017.

\bibitem[Cao et~al.(2023)Cao, Bhupathiraju, Naghavi, Sugawara, Mao, and Rampazzi]{cao2023you}
Yulong Cao, S~Hrushikesh Bhupathiraju, Pirouz Naghavi, Takeshi Sugawara, Z~Morley Mao, and Sara Rampazzi.
\newblock You can't see me: Physical removal attacks on $\{$lidar-based$\}$ autonomous vehicles driving frameworks.
\newblock In \emph{32nd USENIX Security Symposium (USENIX Security 23)}, pages 2993--3010, 2023.

\bibitem[Carlini and Wagner(2017)]{carlini2017cw}
Nicholas Carlini and David Wagner.
\newblock Towards evaluating the robustness of neural networks.
\newblock In \emph{2017 ieee symposium on security and privacy (sp)}, pages 39--57. Ieee, 2017.

\bibitem[Deng et~al.(2020)Deng, Zheng, Zhang, Chen, Lou, and Kim]{deng2020analysis}
Yao Deng, Xi Zheng, Tianyi Zhang, Chen Chen, Guannan Lou, and Miryung Kim.
\newblock An analysis of adversarial attacks and defenses on autonomous driving models.
\newblock In \emph{2020 IEEE international conference on pervasive computing and communications (PerCom)}, pages 1--10. IEEE, 2020.

\bibitem[Dosovitskiy et~al.(2017)Dosovitskiy, Ros, Codevilla, Lopez, and Koltun]{dosovitskiy2017carla}
Alexey Dosovitskiy, German Ros, Felipe Codevilla, Antonio Lopez, and Vladlen Koltun.
\newblock Carla: An open urban driving simulator.
\newblock In \emph{Conference on robot learning}, pages 1--16. PMLR, 2017.

\bibitem[Duan et~al.(2020)Duan, Ma, Wang, Bailey, Qin, and Yang]{duan2020naturalstyles}
Ranjie Duan, Xingjun Ma, Yisen Wang, James Bailey, A~Kai Qin, and Yun Yang.
\newblock Adversarial camouflage: Hiding physical-world attacks with natural styles.
\newblock In \emph{Proceedings of the IEEE/CVF conference on computer vision and pattern recognition}, pages 1000--1008, 2020.

\bibitem[Eykholt et~al.(2018)Eykholt, Evtimov, Fernandes, Li, Rahmati, Xiao, Prakash, Kohno, and Song]{eykholt2018rp2}
Kevin Eykholt, Ivan Evtimov, Earlence Fernandes, Bo Li, Amir Rahmati, Chaowei Xiao, Atul Prakash, Tadayoshi Kohno, and Dawn Song.
\newblock Robust physical-world attacks on deep learning visual classification.
\newblock In \emph{Proceedings of the IEEE conference on computer vision and pattern recognition}, pages 1625--1634, 2018.

\bibitem[Fan et~al.(2023)Fan, Ji, Xu, Cheng, Sakaridis, and Van~Gool]{fan2023advances}
Deng-Ping Fan, Ge-Peng Ji, Peng Xu, Ming-Ming Cheng, Christos Sakaridis, and Luc Van~Gool.
\newblock Advances in deep concealed scene understanding.
\newblock \emph{Visual Intelligence}, 1\penalty0 (1):\penalty0 16, 2023.

\bibitem[Feng et~al.(2021)Feng, Wu, Zhang, Zhang, and Zhang]{feng2021metaattack}
Weiwei Feng, Baoyuan Wu, Tianzhu Zhang, Yong Zhang, and Yongdong Zhang.
\newblock Meta-attack: Class-agnostic and model-agnostic physical adversarial attack.
\newblock In \emph{Proceedings of the IEEE/CVF international conference on computer vision}, pages 7787--7796, 2021.

\bibitem[Goodfellow et~al.(2014)Goodfellow, Shlens, and Szegedy]{goodfellow2014FGSM}
Ian~J Goodfellow, Jonathon Shlens, and Christian Szegedy.
\newblock Explaining and harnessing adversarial examples.
\newblock \emph{arXiv preprint arXiv:1412.6572}, 2014.

\bibitem[Gu et~al.(2023)Gu, Jia, de~Jorge, Yu, Liu, Ma, Xun, Hu, Khakzar, Li, et~al.]{gu2023survey}
Jindong Gu, Xiaojun Jia, Pau de Jorge, Wenqain Yu, Xinwei Liu, Avery Ma, Yuan Xun, Anjun Hu, Ashkan Khakzar, Zhijiang Li, et~al.
\newblock A survey on transferability of adversarial examples across deep neural networks.
\newblock \emph{arXiv preprint arXiv:2310.17626}, 2023.

\bibitem[Gu{\'e}don and Lepetit(2024)]{guedon2024sugar}
Antoine Gu{\'e}don and Vincent Lepetit.
\newblock Sugar: Surface-aligned gaussian splatting for efficient 3d mesh reconstruction and high-quality mesh rendering.
\newblock In \emph{Proceedings of the IEEE/CVF Conference on Computer Vision and Pattern Recognition}, pages 5354--5363, 2024.

\bibitem[He et~al.(2023)He, Liu, Li, Liang, Li, Jia, and Cao]{he2023generating}
Bangyan He, Jian Liu, Yiming Li, Siyuan Liang, Jingzhi Li, Xiaojun Jia, and Xiaochun Cao.
\newblock Generating transferable 3d adversarial point cloud via random perturbation factorization.
\newblock In \emph{Proceedings of the AAAI Conference on Artificial Intelligence}, pages 764--772, 2023.

\bibitem[He et~al.(2016)He, Zhang, Ren, and Sun]{he2016resnet}
Kaiming He, Xiangyu Zhang, Shaoqing Ren, and Jian Sun.
\newblock Deep residual learning for image recognition.
\newblock In \emph{Proceedings of the IEEE conference on computer vision and pattern recognition}, pages 770--778, 2016.

\bibitem[He et~al.(2017)He, Gkioxari, Doll{\'a}r, and Girshick]{he2017maskrcnn}
Kaiming He, Georgia Gkioxari, Piotr Doll{\'a}r, and Ross Girshick.
\newblock Mask r-cnn.
\newblock In \emph{Proceedings of the IEEE international conference on computer vision}, pages 2961--2969, 2017.

\bibitem[Hu et~al.(2021)Hu, Kung, Tan, Chen, Hua, and Cheng]{hu2021naturalistic}
Yu-Chih-Tuan Hu, Bo-Han Kung, Daniel~Stanley Tan, Jun-Cheng Chen, Kai-Lung Hua, and Wen-Huang Cheng.
\newblock Naturalistic physical adversarial patch for object detectors.
\newblock In \emph{Proceedings of the IEEE/CVF International Conference on Computer Vision}, pages 7848--7857, 2021.

\bibitem[Hu et~al.(2022)Hu, Huang, Zhu, Sun, Zhang, and Hu]{hu2022adversarial}
Zhanhao Hu, Siyuan Huang, Xiaopei Zhu, Fuchun Sun, Bo Zhang, and Xiaolin Hu.
\newblock Adversarial texture for fooling person detectors in the physical world.
\newblock In \emph{Proceedings of the IEEE/CVF conference on computer vision and pattern recognition}, pages 13307--13316, 2022.

\bibitem[Hu et~al.(2023)Hu, Chu, Zhu, Zhang, Zhang, and Hu]{hu2023physically}
Zhanhao Hu, Wenda Chu, Xiaopei Zhu, Hui Zhang, Bo Zhang, and Xiaolin Hu.
\newblock Physically realizable natural-looking clothing textures evade person detectors via 3d modeling.
\newblock In \emph{Proceedings of the IEEE/CVF Conference on Computer Vision and Pattern Recognition}, pages 16975--16984, 2023.

\bibitem[Huang et~al.(2020)Huang, Gao, Zhou, Xie, Yuille, Zou, and Liu]{huang2020universal}
Lifeng Huang, Chengying Gao, Yuyin Zhou, Cihang Xie, Alan~L Yuille, Changqing Zou, and Ning Liu.
\newblock Universal physical camouflage attacks on object detectors.
\newblock In \emph{Proceedings of the IEEE/CVF conference on computer vision and pattern recognition}, pages 720--729, 2020.

\bibitem[Huang et~al.(2024)Huang, Dong, Ruan, Yang, Su, and Wei]{huang2024tt3d}
Yao Huang, Yinpeng Dong, Shouwei Ruan, Xiao Yang, Hang Su, and Xingxing Wei.
\newblock Towards transferable targeted 3d adversarial attack in the physical world.
\newblock In \emph{Proceedings of the IEEE/CVF Conference on Computer Vision and Pattern Recognition}, pages 24512--24522, 2024.

\bibitem[Jia et~al.(2020)Jia, Wei, Cao, and Han]{jia2020adv}
Xiaojun Jia, Xingxing Wei, Xiaochun Cao, and Xiaoguang Han.
\newblock Adv-watermark: A novel watermark perturbation for adversarial examples.
\newblock In \emph{Proceedings of the 28th ACM international conference on multimedia}, pages 1579--1587, 2020.

\bibitem[Jia et~al.(2024)Jia, Gao, Guo, Ma, Huang, Qin, Liu, and Cao]{jia2024semantic}
Xiaojun Jia, Sensen Gao, Qing Guo, Ke Ma, Yihao Huang, Simeng Qin, Yang Liu, and Xiaochun Cao.
\newblock Semantic-aligned adversarial evolution triangle for high-transferability vision-language attack.
\newblock \emph{arXiv preprint arXiv:2411.02669}, 2024.

\bibitem[Jia et~al.(2025)Jia, Gao, Qin, Pang, Du, Huang, Li, Li, Li, and Liu]{jia2025adversarial}
Xiaojun Jia, Sensen Gao, Simeng Qin, Tianyu Pang, Chao Du, Yihao Huang, Xinfeng Li, Yiming Li, Bo Li, and Yang Liu.
\newblock Adversarial attacks against closed-source mllms via feature optimal alignment.
\newblock \emph{arXiv preprint arXiv:2505.21494}, 2025.

\bibitem[Kato et~al.(2018)Kato, Ushiku, and Harada]{Kato2018NMR}
Hiroharu Kato, Yoshitaka Ushiku, and Tatsuya Harada.
\newblock Neural 3d mesh renderer.
\newblock In \emph{Proceedings of the IEEE Conference on Computer Vision and Pattern Recognition (CVPR)}, 2018.

\bibitem[Kerbl et~al.(2023)Kerbl, Kopanas, Leimk{\"u}hler, and Drettakis]{kerbl20233dgs}
Bernhard Kerbl, Georgios Kopanas, Thomas Leimk{\"u}hler, and George Drettakis.
\newblock 3d gaussian splatting for real-time radiance field rendering.
\newblock \emph{ACM Trans. Graph.}, 42\penalty0 (4):\penalty0 139--1, 2023.

\bibitem[Kirillov et~al.(2023)Kirillov, Mintun, Ravi, Mao, Rolland, Gustafson, Xiao, Whitehead, Berg, Lo, et~al.]{kirillov2023SAM}
Alexander Kirillov, Eric Mintun, Nikhila Ravi, Hanzi Mao, Chloe Rolland, Laura Gustafson, Tete Xiao, Spencer Whitehead, Alexander~C Berg, Wan-Yen Lo, et~al.
\newblock Segment anything.
\newblock In \emph{Proceedings of the IEEE/CVF International Conference on Computer Vision}, pages 4015--4026, 2023.

\bibitem[Kong et~al.(2024{\natexlab{a}})Kong, Liang, and Ren]{kong2024environmental}
Dehong Kong, Siyuan Liang, and Wenqi Ren.
\newblock Environmental matching attack against unmanned aerial vehicles object detection.
\newblock \emph{arXiv preprint arXiv:2405.07595}, 2024{\natexlab{a}}.

\bibitem[Kong et~al.(2024{\natexlab{b}})Kong, Liang, Zhu, Zhong, and Ren]{kong2024patch}
Dehong Kong, Siyuan Liang, Xiaopeng Zhu, Yuansheng Zhong, and Wenqi Ren.
\newblock Patch is enough: naturalistic adversarial patch against vision-language pre-training models.
\newblock \emph{Visual Intelligence}, 2\penalty0 (1):\penalty0 1--10, 2024{\natexlab{b}}.

\bibitem[Kurakin et~al.(2018)Kurakin, Goodfellow, and Bengio]{kurakin2018ifgsm}
Alexey Kurakin, Ian~J Goodfellow, and Samy Bengio.
\newblock Adversarial examples in the physical world.
\newblock In \emph{Artificial intelligence safety and security}, pages 99--112. Chapman and Hall/CRC, 2018.

\bibitem[Li et~al.(2023)Li, Lian, and Chen]{li2023adv3d}
Leheng Li, Qing Lian, and Ying-Cong Chen.
\newblock Adv3d: generating 3d adversarial examples in driving scenarios with nerf.
\newblock \emph{arXiv preprint arXiv:2309.01351}, 2023.

\bibitem[Lian et~al.(2022)Lian, Mei, Zhang, and Ma]{lian2022benchmarking}
Jiawei Lian, Shaohui Mei, Shun Zhang, and Mingyang Ma.
\newblock Benchmarking adversarial patch against aerial detection.
\newblock \emph{IEEE Transactions on Geoscience and Remote Sensing}, 60:\penalty0 1--16, 2022.

\bibitem[Liang et~al.(2020)Liang, Wei, Yao, and Cao]{liang2020efficient}
Siyuan Liang, Xingxing Wei, Siyuan Yao, and Xiaochun Cao.
\newblock Efficient adversarial attacks for visual object tracking.
\newblock In \emph{Computer Vision--ECCV 2020: 16th European Conference, Glasgow, UK, August 23--28, 2020, Proceedings, Part XXVI 16}, 2020.

\bibitem[Liang et~al.(2021)Liang, Wei, and Cao]{liang2021generate}
Siyuan Liang, Xingxing Wei, and Xiaochun Cao.
\newblock Generate more imperceptible adversarial examples for object detection.
\newblock In \emph{ICML 2021 Workshop on Adversarial Machine Learning}, 2021.

\bibitem[Liang et~al.(2022{\natexlab{a}})Liang, Li, Fan, Jia, Li, Wu, and Cao]{liang2022large}
Siyuan Liang, Longkang Li, Yanbo Fan, Xiaojun Jia, Jingzhi Li, Baoyuan Wu, and Xiaochun Cao.
\newblock A large-scale multiple-objective method for black-box attack against object detection.
\newblock In \emph{European Conference on Computer Vision}, 2022{\natexlab{a}}.

\bibitem[Liang et~al.(2022{\natexlab{b}})Liang, Wu, Fan, Wei, and Cao]{liang2022parallel}
Siyuan Liang, Baoyuan Wu, Yanbo Fan, Xingxing Wei, and Xiaochun Cao.
\newblock Parallel rectangle flip attack: A query-based black-box attack against object detection.
\newblock \emph{arXiv preprint arXiv:2201.08970}, 2022{\natexlab{b}}.

\bibitem[Liang et~al.(2024)Liang, Wang, Chen, Liu, Wu, Chang, Cao, and Tao]{liang2024object}
Siyuan Liang, Wei Wang, Ruoyu Chen, Aishan Liu, Boxi Wu, Ee-Chien Chang, Xiaochun Cao, and Dacheng Tao.
\newblock Object detectors in the open environment: Challenges, solutions, and outlook.
\newblock \emph{arXiv preprint arXiv:2403.16271}, 2024.

\bibitem[Liu et~al.(2023)Liu, Guo, Wang, Liang, Tao, Zhou, Liu, Liu, and Tao]{liu2023x}
Aishan Liu, Jun Guo, Jiakai Wang, Siyuan Liang, Renshuai Tao, Wenbo Zhou, Cong Liu, Xianglong Liu, and Dacheng Tao.
\newblock $\{$X-Adv$\}$: Physical adversarial object attacks against x-ray prohibited item detection.
\newblock In \emph{32nd USENIX Security Symposium (USENIX Security 23)}, 2023.

\bibitem[Lou et~al.(2024)Lou, Jia, Gu, Liu, Liang, He, and Cao]{lou2024hitadv}
Tianrui Lou, Xiaojun Jia, Jindong Gu, Li Liu, Siyuan Liang, Bangyan He, and Xiaochun Cao.
\newblock Hide in thicket: Generating imperceptible and rational adversarial perturbations on 3d point clouds.
\newblock In \emph{Proceedings of the IEEE/CVF Conference on Computer Vision and Pattern Recognition}, pages 24326--24335, 2024.

\bibitem[Mildenhall et~al.(2021)Mildenhall, Srinivasan, Tancik, Barron, Ramamoorthi, and Ng]{mildenhall2021nerf}
Ben Mildenhall, Pratul~P Srinivasan, Matthew Tancik, Jonathan~T Barron, Ravi Ramamoorthi, and Ren Ng.
\newblock Nerf: Representing scenes as neural radiance fields for view synthesis.
\newblock \emph{Communications of the ACM}, 65\penalty0 (1):\penalty0 99--106, 2021.

\bibitem[Moosavi-Dezfooli et~al.(2017)Moosavi-Dezfooli, Fawzi, Fawzi, and Frossard]{moosavi2017uap}
Seyed-Mohsen Moosavi-Dezfooli, Alhussein Fawzi, Omar Fawzi, and Pascal Frossard.
\newblock Universal adversarial perturbations.
\newblock In \emph{Proceedings of the IEEE conference on computer vision and pattern recognition}, pages 1765--1773, 2017.

\bibitem[Muxue et~al.()Muxue, Wang, Liang, Liu, Liu, Yang, and Cao]{muxue2023adversarial}
Liang Muxue, Chuan Wang, Siyuan Liang, Aishan Liu, Zeming Liu, Liang Yang, and Xiaochun Cao.
\newblock Adversarial instance attacks for interactions between human and object.

\bibitem[Nguyen et~al.(2023)Nguyen, Fernando, Fookes, and Sridharan]{nguyen2023physical}
Kien Nguyen, Tharindu Fernando, Clinton Fookes, and Sridha Sridharan.
\newblock Physical adversarial attacks for surveillance: A survey.
\newblock \emph{IEEE Transactions on Neural Networks and Learning Systems}, 2023.

\bibitem[Ren et~al.(2015)Ren, He, Girshick, and Sun]{ren2015frcnn}
Shaoqing Ren, Kaiming He, Ross Girshick, and Jian Sun.
\newblock Faster r-cnn: Towards real-time object detection with region proposal networks.
\newblock \emph{Advances in neural information processing systems}, 28, 2015.

\bibitem[Sharif et~al.(2016)Sharif, Bhagavatula, Bauer, and Reiter]{sharif2016nps}
Mahmood Sharif, Sruti Bhagavatula, Lujo Bauer, and Michael~K Reiter.
\newblock Accessorize to a crime: Real and stealthy attacks on state-of-the-art face recognition.
\newblock In \emph{Proceedings of the 2016 acm sigsac conference on computer and communications security}, pages 1528--1540, 2016.

\bibitem[Snavely et~al.(2006)Snavely, Seitz, and Szeliski]{noah2006sfm}
Noah Snavely, Steven~M. Seitz, and Richard Szeliski.
\newblock Photo tourism.
\newblock \emph{ACM Transactions on Graphics}, page 835–846, 2006.

\bibitem[Song et~al.(2018)Song, Eykholt, Evtimov, Fernandes, Li, Rahmati, Tramer, Prakash, and Kohno]{song2018physical_objectdetection}
Dawn Song, Kevin Eykholt, Ivan Evtimov, Earlence Fernandes, Bo Li, Amir Rahmati, Florian Tramer, Atul Prakash, and Tadayoshi Kohno.
\newblock Physical adversarial examples for object detectors.
\newblock In \emph{12th USENIX workshop on offensive technologies (WOOT 18)}, 2018.

\bibitem[Sun et~al.(2023)Sun, Yao, Jiang, Wang, and Chen]{sun2023differential}
Jialiang Sun, Wen Yao, Tingsong Jiang, Donghua Wang, and Xiaoqian Chen.
\newblock Differential evolution based dual adversarial camouflage: Fooling human eyes and object detectors.
\newblock \emph{Neural Networks}, 163:\penalty0 256--271, 2023.

\bibitem[Suryanto et~al.(2022)Suryanto, Kim, Kang, Larasati, Yun, Le, Yang, Oh, and Kim]{suryanto2022DTA}
Naufal Suryanto, Yongsu Kim, Hyoeun Kang, Harashta~Tatimma Larasati, Youngyeo Yun, Thi-Thu-Huong Le, Hunmin Yang, Se-Yoon Oh, and Howon Kim.
\newblock Dta: Physical camouflage attacks using differentiable transformation network.
\newblock In \emph{Proceedings of the IEEE/CVF Conference on Computer Vision and Pattern Recognition}, pages 15305--15314, 2022.

\bibitem[Suryanto et~al.(2023)Suryanto, Kim, Larasati, Kang, Le, Hong, Yang, Oh, and Kim]{suryanto2023ACTIVE}
Naufal Suryanto, Yongsu Kim, Harashta~Tatimma Larasati, Hyoeun Kang, Thi-Thu-Huong Le, Yoonyoung Hong, Hunmin Yang, Se-Yoon Oh, and Howon Kim.
\newblock Active: Towards highly transferable 3d physical camouflage for universal and robust vehicle evasion.
\newblock In \emph{Proceedings of the IEEE/CVF International Conference on Computer Vision}, pages 4305--4314, 2023.

\bibitem[Thys et~al.(2019)Thys, Van~Ranst, and Goedem{\'e}]{thys2019fooling}
Simen Thys, Wiebe Van~Ranst, and Toon Goedem{\'e}.
\newblock Fooling automated surveillance cameras: adversarial patches to attack person detection.
\newblock In \emph{Proceedings of the IEEE/CVF conference on computer vision and pattern recognition workshops}, pages 0--0, 2019.

\bibitem[Vaswani(2017)]{vaswani2017attention}
A Vaswani.
\newblock Attention is all you need.
\newblock \emph{Advances in Neural Information Processing Systems}, 2017.

\bibitem[Wang et~al.(2022)Wang, Jiang, Sun, Zhou, Gong, Zhang, Yao, and Chen]{wang2022FCA}
Donghua Wang, Tingsong Jiang, Jialiang Sun, Weien Zhou, Zhiqiang Gong, Xiaoya Zhang, Wen Yao, and Xiaoqian Chen.
\newblock Fca: Learning a 3d full-coverage vehicle camouflage for multi-view physical adversarial attack.
\newblock In \emph{Proceedings of the AAAI conference on artificial intelligence}, pages 2414--2422, 2022.

\bibitem[Wang et~al.(2021)Wang, Liu, Yin, Liu, Tang, and Liu]{wang2021DAS}
Jiakai Wang, Aishan Liu, Zixin Yin, Shunchang Liu, Shiyu Tang, and Xianglong Liu.
\newblock Dual attention suppression attack: Generate adversarial camouflage in physical world.
\newblock In \emph{Proceedings of the IEEE/CVF conference on computer vision and pattern recognition}, pages 8565--8574, 2021.

\bibitem[Wang et~al.(2024{\natexlab{a}})Wang, Liu, Yin, Wang, Guo, Qin, Wu, and Liu]{wang2024TAS}
Jiakai Wang, Xianglong Liu, Zixin Yin, Yuxuan Wang, Jun Guo, Haotong Qin, Qingtao Wu, and Aishan Liu.
\newblock Generate transferable adversarial physical camouflages via triplet attention suppression.
\newblock \emph{International Journal of Computer Vision}, pages 1--17, 2024{\natexlab{a}}.

\bibitem[Wang et~al.(2023)Wang, Luo, Sato, Xu, and Chen]{wang2023does}
Ningfei Wang, Yunpeng Luo, Takami Sato, Kaidi Xu, and Qi~Alfred Chen.
\newblock Does physical adversarial example really matter to autonomous driving? towards system-level effect of adversarial object evasion attack.
\newblock In \emph{Proceedings of the IEEE/CVF International Conference on Computer Vision}, pages 4412--4423, 2023.

\bibitem[Wang et~al.(2024{\natexlab{b}})Wang, Mei, Lian, and Lu]{wang2024fooling}
Xiaofei Wang, Shaohui Mei, Jiawei Lian, and Yingjie Lu.
\newblock Fooling aerial detectors by background attack via dual-adversarial-induced error identification.
\newblock \emph{IEEE Transactions on Geoscience and Remote Sensing}, 2024{\natexlab{b}}.

\bibitem[Wang et~al.(2019)Wang, Zheng, Song, Wang, Rahimpour, and Qi]{wang2019advpattern}
Zhibo Wang, Siyan Zheng, Mengkai Song, Qian Wang, Alireza Rahimpour, and Hairong Qi.
\newblock advpattern: Physical-world attacks on deep person re-identification via adversarially transformable patterns.
\newblock In \emph{Proceedings of the IEEE/CVF International Conference on Computer Vision}, pages 8341--8350, 2019.

\bibitem[Wei et~al.(2018)Wei, Liang, Chen, and Cao]{wei2018transferable}
Xingxing Wei, Siyuan Liang, Ning Chen, and Xiaochun Cao.
\newblock Transferable adversarial attacks for image and video object detection.
\newblock \emph{arXiv preprint arXiv:1811.12641}, 2018.

\bibitem[Wu et~al.(2020)Wu, Ning, Li, Huang, Yang, and Wang]{wu2020genetic}
Tong Wu, Xuefei Ning, Wenshuo Li, Ranran Huang, Huazhong Yang, and Yu Wang.
\newblock Physical adversarial attack on vehicle detector in the carla simulator.
\newblock \emph{arXiv preprint arXiv:2007.16118}, 2020.

\bibitem[Xu et~al.(2020)Xu, Zhang, Liu, Fan, Sun, Chen, Chen, Wang, and Lin]{xu2020adversarial}
Kaidi Xu, Gaoyuan Zhang, Sijia Liu, Quanfu Fan, Mengshu Sun, Hongge Chen, Pin-Yu Chen, Yanzhi Wang, and Xue Lin.
\newblock Adversarial t-shirt! evading person detectors in a physical world.
\newblock In \emph{Computer Vision--ECCV 2020: 16th European Conference, Glasgow, UK, August 23--28, 2020, Proceedings, Part V 16}, pages 665--681. Springer, 2020.

\bibitem[Zhang et~al.(2018)Zhang, Foroosh, David, and Gong]{zhang2018camou}
Yang Zhang, Hassan Foroosh, Philip David, and Boqing Gong.
\newblock Camou: Learning physical vehicle camouflages to adversarially attack detectors in the wild.
\newblock In \emph{International Conference on Learning Representations}, 2018.

\bibitem[Zhang et~al.(2023)Zhang, Gong, Zhang, Bin, Li, Qi, Wen, and Zhong]{zhang2023boosting}
Yu Zhang, Zhiqiang Gong, Yichuang Zhang, Kangcheng Bin, Yongqian Li, Jiahao Qi, Hao Wen, and Ping Zhong.
\newblock Boosting transferability of physical attack against detectors by redistributing separable attention.
\newblock \emph{Pattern Recognition}, 138:\penalty0 109435, 2023.

\bibitem[Zhou et~al.(2024)Zhou, Lyu, He, and Li]{zhou2024rauca}
Jiawei Zhou, Linye Lyu, Daojing He, and Yu Li.
\newblock Rauca: A novel physical adversarial attack on vehicle detectors via robust and accurate camouflage generation.
\newblock \emph{arXiv preprint arXiv:2402.15853}, 2024.

\bibitem[Zhu and Rong(2024)]{zhu2024multiview}
Heran Zhu and Dazhong Rong.
\newblock Multiview consistent physical adversarial camouflage generation through semantic guidance.
\newblock In \emph{2024 International Joint Conference on Neural Networks (IJCNN)}, pages 1--8. IEEE, 2024.

\bibitem[Zhu et~al.(2020)Zhu, Su, Lu, Li, Wang, and Dai]{zhu2020ddetr}
Xizhou Zhu, Weijie Su, Lewei Lu, Bin Li, Xiaogang Wang, and Jifeng Dai.
\newblock Deformable detr: Deformable transformers for end-to-end object detection.
\newblock \emph{arXiv preprint arXiv:2010.04159}, 2020.

\end{thebibliography}
